\documentclass[12pt,pra,aps,amssymb,amsfonts,amsmath,tightenlines]{revtex4}
\usepackage[caption = false]{subfig}
\usepackage{dsfont}
\usepackage{graphicx}
\usepackage{amssymb,amsfonts,amsthm}
\usepackage{color}
\usepackage{bm,float}

\begin{document}

\title{Yes, Prime Minister, question order \textit {does} matter -- and it's certainly not classical! 
But is it quantum?}
\author{Dorje~C.~Brody}

\affiliation{
School of Mathematics and Physics, University of Surrey, 
Guildford GU2 7XH, UK 
}

\date{\today}
\begin{abstract}
\noindent 
Response to a poll can be manipulated by means of a series of leading questions. We show that such phenomena cannot be explained by use of classical probability theory, whereas quantum probability theory admits a possibility of offering an explanation. Admissible transformation rules in quantum probability, however, do impose some constraints on the modelling of cognitive behaviour, which are highlighted here. Focusing on a recent poll conducted by Ipsos on a set of questions posed by Sir Humphrey Appleby in an episode of the British political satire \textit{Yes, Prime Minister}, we show that the resulting data cannot be explained quite so simply using quantum rules, although it seems not impossible. 
\vspace{-0.2cm}
\\
\end{abstract}

\maketitle

\section*{Introduction} 

In an episode of the satirical British political sitcom \textit{Yes, Prime Minister} from the 
1980s, Sir Humphrey Appleby once explained to Bernard Woolley (two of the characters) 
how it is possible to get contradictory polling results by asking a series of leading questions 
beforehand. The polling discussed in the episode concerns whether the public is for or 
against the reintroduction of national service. Recently, the leading questions 
outlined by Appleby were put to the public by the market research and polling giant Ipsos, 
the findings of which have been made public to raise awareness of the fact that people can be 
misled by means of a such questions \cite{Ipsos}. 
The actual experiment conducted by Ipsos is explained 
on their web site: ``Ipsos interviewed a representative quota sample of 2,158 adults 
aged 16-75 in Great Britain. Half saw the `Sample A' questions, reflecting a positive view 
about national service. Half saw `Sample B', reflecting a negative view.'' For the convenience 
of readers I quote their findings in Tables \ref{Table:1} and \ref{Table:2} below. 

The leading questions are there to provide contexts to the final question, and to many there 
is little surprise in the results indicated in the two tables. The issue, however, is that such 
context-dependent outcomes are not compatible with laws of classical probability. Leaving 
aside the minor semantic debate in Question 5 of the two samples (one could reasonably 
argue that not opposing need not be the same as favouring), in one case the data shows 
that 45\% of the people are in support of reintroducing national service as opposed to 
38\% against, while in the other case this figure is 34\% for and 48\% against -- thus 
violating the law of conditional probability. The argument for the violation goes as 
follows. Let $A$ denote the series of four leading questions in Sample A. Each question can 
be answered in three different ways, so in total there are 12 possible ways $\{A_i\}$ to 
answer these questions. Similarly, let $B$ denote the series of four questions in Sample B, 
which can again be answered in 12 possible ways $\{B_i\}$. Let $X$ be the fifth question, on 
national service, which we assume to be identical in both samples. Then we have 
\[
{\mathbb P}(X={\rm Yes}) = 
\sum_{i=1}^{12} {\mathbb P}(X={\rm Yes}|A=A_i) {\mathbb P}(A=A_i)= 
\sum_{i=1}^{12} {\mathbb P}(X={\rm Yes}|B=B_i) {\mathbb P}(B=B_i)
\] 
from the law of conditional probability. Yet, the results in the two Tables suggest otherwise. 
The phenomenon seems intuitively understandable, but classical probability theory fails to 
adequately model empirical data here.

\begin{table}[ht]
    \centering
    \begin{tabular}{p{0.53\linewidth} | p{0.09\linewidth} | p{0.09\linewidth} | p{0.09\linewidth}}
      Sample A &  ~Yes &  ~Unsure &  ~No \\ \hline
      1. Are you worried about the number of young people without jobs?  
      &  ~52 \% & ~9\% & ~39\%   \\ \hline
      2. Are you worried about the rise in crime among teenagers? 
      &  ~81\% & ~4\% & ~15\%   \\ \hline
      3. Do you think there’s a lack of discipline in Britain’s comprehensive schools? 
      &  ~72\% & ~13\% & ~15\% \\  \hline
      4. Do you think young people would welcome some authority and leadership in their lives? 
      &  ~59\% & ~16\% & ~25\% \\  \hline
      5. \textbf{Would you be in \underline{favour} of reintroducing National Service in Britain?} 
      &  ~45\% & ~17\% & ~38\%    \\
    \end{tabular}
    \caption{\textit{First set of leading questions}. The data, obtained by Ipsos \cite{Ipsos}, 
    shows that if the question is leading towards a positive view on the reintroduction of 
    national service, then 45\% of people are in support of the idea, whereas 38\% 
    are against. }
    \label{Table:1}
\end{table}

\begin{table}[ht]
    \centering
    \begin{tabular}{p{0.53\linewidth} | p{0.09\linewidth} | p{0.09\linewidth} | p{0.09\linewidth}}
      Sample B &  ~Yes &  ~Unsure & ~No \\ \hline
      1. Are you worried about the danger of war?  
      &  ~63 \% & ~7\% & ~30\%   \\ \hline
      2. Are you worried about the growth of armaments / weapons around the world? 
     &  ~73\% & ~5\% & ~22\%   \\ \hline
     3. Do you think there’s a danger in giving young people guns and teaching them how to kill?
     &  ~79\% & ~6\% & ~15\% \\  \hline
     4. Do you think it’s wrong to force people to take up arms against their will?  
     &  ~79\% & ~9\% & ~12\% \\  \hline
     5. \textbf{Would you \underline{oppose} the reintroduction of National Service in Britain?} 
     &  ~48\% & ~18\% & ~34\%    \\
    \end{tabular}
    \caption{\textit{Second set of leading questions}. The data, obtained by Ipsos \cite{Ipsos}, 
    shows that if the question is leading towards a negative view on the reintroduction of 
    national service, then 34\% of people are in support of the idea, whereas 48\% 
    are against.} 
    \label{Table:2}
\end{table}

In cognitive psychology, such violations is commonly observed. One often-cited example is 
that due to Moore \cite{Moore}. In essence it goes as follows. When people were asked if they 
thought President Bill Clinton was honest, 50\% answered ``Yes''. When they were 
subsequently asked if they thought his Vice President Al Gore was honest, 57\% answered 
``Yes''. However, if the order of the questions were reversed, then the answers were 
68\% ``Yes'' for Gore and 60\% ``Yes'' for Clinton, again contradicting rules of classical 
probability. Countless such examples have led cognitive scientist J.~R.~Busemeyer and many 
others \cite{Busemeyer3,Suppes,Franco,Sornette,Busemeyer0,Busemeyer1,Busemeyer5,
Busemeyer6,Khrennikov,Busemeyer4,Sornette2,Busemeyer2,Ozawa}
to observe that context-dependent probabilistic outcomes can be explained using 
the rules of quantum probability -- an observation that led to the development of a new field 
now known as quantum cognition. 

Quantum cognition is based on the idea that when people are in doubt about making a 
choice from a set of alternatives, then the state of mind associated to that choice is in 
a coherent superposition 
of the different alternatives. Here, a state of mind is defined as the probability 
distribution for choosing the alternatives and these probabilities can be modelled 
on Hilbert space by use of the square-root map \cite{Brody2023}. Working on a Hilbert space to 
analyze probability and statistics, a given distribution can be modelled either as a 
coherent superposition (attributed as a quantum feature) or as a mixed-state density 
matrix (attributed as a classical feature). If the observable quantities (random variables) that 
generate data are commutative, then the two representations (a pure state or its 
associated mixed state) are indistinguishable. However, in cognitive science there are 
numerous examples that show that propositions (statements, questions, etc.) are not 
compatible \cite{Busemeyer2}, and hence it is possible to distinguish pure states from their 
mixed counterparts. A range of paradoxical cognitive behavioural phenomena -- paradoxical from the 
perspective of classical probability theory -- can then be explained consistently by use of the 
mathematical formalism of quantum theory. 

With this in mind, the purpose of the present paper is to highlight the fact that the results 
of Ipsos study listed in the two tables, although incompatible with rules of classical probability, 
nonetheless cannot be explained quite so straightforwardly by standard formulations in 
quantum cognition. First I take up the famous example of 
the conjunction fallacy in cognitive science and illustrate how the phenomenon can 
be explained by means of quantum probability, for the benefit of readers less acquainted 
with the ideas in quantum cognition. The setup for the conjunction fallacy is then 
modified slightly to show that the effect resulting from a leading question can easily be 
explained in this quantum context. 

I then examine the Ipsos data on Appleby's questions and show that this data cannot be explained 
by use of the standard formalism of quantum cognition whereby questions are modelled by 
a set of incompatible observables acting on the Hilbert space. This is followed by the 
consideration of using entangled states to attempt to model the data. After clarifying some 
differences between classical and quantum probabilities, I show that the leading-question 
effect cannot be explained using an entangled state along with local observables. I then 
demonstrate a particular scheme in which the relevant data can be modelled, although the 
validity of the assumptions underlying this scheme can only be established with new 
experimental data. The paper concludes with a discussion on quantum probability rules 
and rational behaviour.

\section*{Conjunction fallacy revisited} 

One key ingredient in quantum probability that distinguishes it from its classical counterpart 
is that questions (observables) are in general not compatible, and hence question order \textit{does}
matter. Perhaps the simplest way of seeing this is to consider a spin measurement of a 
spin-$\frac{1}{2}$ particle. Let $X$ denote the question whether the measurement of the 
spin ${\hat\sigma}_x$ in the $x$ direction is up, and similarly $Y$ for the spin measurement 
in the $y$ direction. Suppose that the initial state of the particle is the ${\hat\sigma}_x$ 
eigenstate with the eigenvalue $+1$ (that is, $X={\rm UP}$ state). Then clearly we have 
\[ 
{\mathbb P}(X={\rm UP}) = 1 . 
\] 
However, if we first measure ${\hat\sigma}_y$, and subsequently measure ${\hat\sigma}_x$, 
then irrespective of the outcome of the $Y$ measurement we have 
\[ 
{\mathbb P}(X={\rm UP}) = \tfrac{1}{2} . 
\] 
If the $X$ and $Y$ measurements were to model two binary questions, we therefore find 
that in general question-order effects (that the answers depend on the order of questions) 
are consequences of incompatible (non-commuting) questions. This is the quintessential 
argument that the mathematical structures underpinning quantum theory leading to probability 
assignment rules are more adequate for modelling human cognitive behaviour than their 
classical counterparts. 

With this in mind, let us discuss the case of the conjunction fallacy \cite{KT}. 
The most well-publicized example of a conjunction fallacy concerns an experiment 
in which participants are first given information about the profile of a hypothetical person 
Linda, that they are single, outspoken, bright, concerned about social justice, and so on. 
The profile is designed to offer the impression that Linda is more likely to be a feminist 
than being a bank teller. Participants are then asked to rank-order the likelihoods of 
Linda being a bank teller ($B$), a bank teller and a feminist ($B$ \& $F$), and a feminist 
($F$). Perhaps not surprisingly, the findings were that 
\[ 
{\mathbb P}(F) > {\mathbb P}(B\,\&\,F) > {\mathbb P}(B). 
\]
In classical probability, while the first inequality is admissible, the second one is problematic 
because it cannot be that the probability of Linda being both a feminist and a bank teller is 
larger than that of being merely a bank teller. 

In the literature of quantum cognition it is understood that the constraint of classical probability 
here can be circumvented on account of the assumption that measurement of $B$ and measurement of 
$F$ are not compatible. The analysis presented, for instance, in \cite{Franco,Busemeyer2} 
shows how this can be explained. Their argument goes in essence as follows. 
If we let $q$ denote the probability that Linda is a 
bank teller and let $p$ denote the probability that Linda is a feminist, then the initial state can be 
expressed in two different ways: 
\[ 
|\psi\rangle = \sqrt{p}\,|F\rangle + \sqrt{1-p}\,|{\bar F}\rangle = 
\sqrt{q}\,|B\rangle + \sqrt{1-q}\,|{\bar B}\rangle . 
\]
Here, $|F\rangle$ denotes the state that Linda is a feminist, 
$|{\bar F}\rangle=|\neg F\rangle$ denotes the state that Linda is not a feminist, and 
similarly for $|B\rangle$ and $|{\bar B}\rangle$ for the states of being a bank teller or not. The 
argument of \cite{Busemeyer2} is that in assessing the probability 
${\mathbb P}(B\,\&\,F)$, it is natural to first to ask whether Linda is a feminist because of 
the association with the provided profile of Linda. Once the result of the feminist question 
is affirmative, and if $|B\rangle$ is closer to $|F\rangle$ than to $|\psi\rangle$, then the 
likelihood of $B$ conditional on $F$ becomes larger than the initial likelihood of $B$. In 
other words, the state $|\psi\rangle$ can be such that we obtain the inequality 
$p>p\,|\langle B|F\rangle|^2 > q$ to explain the data. 

This experiment can be modified slightly into the following thought experiment 
to assess the impact of leading questioners. 
Once the profile of Linda is provided, participants are split into two groups. One group 
of participants is asked if Linda is a bank teller, which will determine ${\mathbb P}(B)$; 
another group is first asked if she is a feminist, which will determine ${\mathbb P}(F)$, 
and this is followed by the question if they thought she could be a bank teller. This latter 
question will then 
determine the likelihood ${\mathbb P}^F(B)$ that she is thought to be a bank teller, having 
being assessed whether she might be a feminist. Importantly, though, this is not a 
conditional statement about the answer to the first question (as in the conjunction 
fallacy experiment), because irrespective of the answer to the feminist question, all 
participants in the second group are asked the followup question. 
If it turns out that ${\mathbb P}^F(B)\neq 
{\mathbb P}(B)$, then we can interpret the feminist question as a leading question to 
alter the outcome of the bank-teller question. (This experiment, in a different setup, has 
indeed been performed to show this effect \cite{Busemeyer6}.) 

\begin{figure}[t]
  \centering
       {\includegraphics[width=0.680\textwidth]{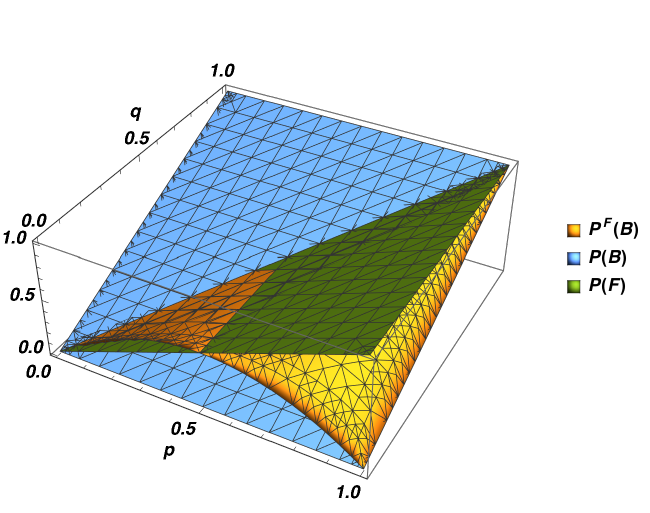}}\hfill
\caption{\textit{Probability comparison}. The probability ${\mathbb P}^F(B)$ that Linda 
is a bank teller, after one has assessed whether she might be a feminist, is compared here 
with the probability ${\mathbb P}(B)=q$ that she is thought to be a bank teller, and the 
probability ${\mathbb P}(F)=p$ that she is thought to be a feminist. In the triangular 
region $0<q<p<1$, we have ${\mathbb P}^F(B)>{\mathbb P}(B)$, indicating that the 
feminist question can be used as a leading question to enhance the assessment that 
Linda might be a bank teller. 
} 
\label{fig:1} 
\end{figure}

To work out the conditional probability ${\mathbb P}^F(B)$, it suffices to consider the 
state after the feminist question is settled. This is given by the density matrix 
\[ 
{\hat\rho}_F = p \, |F\rangle\langle F| + (1-p) \, |{\bar F}\rangle\langle{\bar F}| . 
\] 
Then we have ${\mathbb P}^F(B) = \langle B|{\hat\rho}_F|B\rangle$. To calculate this 
expectation, it helps to express $|B\rangle$ in terms of $|F\rangle$ and 
$|{\bar F}\rangle$. Writing $|B\rangle=\sqrt{\alpha}|F\rangle+\sqrt{1-\alpha}|{\bar F}\rangle$, 
a short calculation shows that 
\[
\sqrt{\alpha}=\sqrt{pq}+\sqrt{(1-p)(1-q)} \, , 
\]
from which we deduce that 
\[ 
{\mathbb P}^F(B) = 
2p(p-1)(2q-1)+q+2(2p-1)\sqrt{pq(1-p)(1-q)} \, . 
\] 
In Figure~\ref{fig:1} the three probabilities ${\mathbb P}(B)=q$, ${\mathbb P}(F)=p$, 
and ${\mathbb P}^F(B)$ are plotted as functions of $p$ and $q$. The figure clearly 
shows that there is a large region in parameter space for which the inequality 
\[ 
{\mathbb P}(F) > {\mathbb P}^F(B) > {\mathbb P}(B) 
\]
holds. In fact, the equality ${\mathbb P}^F(B)={\mathbb P}(B)$ expected to follow from 
classical probability assignment rules holds if and only if $p=q$. But this is precisely the 
case in which the two observables $F$ and $B$ commute, i.e. when they are compatible. 
In fact, ${\mathbb P}^F(B)={\mathbb P}(B)$ also holds if $p=0$ or $p=1$, but these 
cases are redundant because the initial state is already an eigenstate of $F$ and hence 
measurement of $F$ will have no impact. It follows that phenomena of the type 
${\mathbb P}^F(B)\neq {\mathbb P}(B)$ considered here (often referred to as the 
interference effect in the quantum cognition literature because measurement of $F$ 
interferes with the outcome of $B$), which are commonly observed in psychology experiments, 
demand the application of quantum probability theory. 

\section*{Appleby's leading questions} 

Returning to the effect of leading-questions shown on Tables \ref{Table:1} and \ref{Table:2}, 
in line with the standard formulation in quantum cognition whereby asking a question is 
modelled as performing a measurement of an observable, it would be interesting to see 
whether the data obtained by Ipsos can be modelled by means of a suitably chosen initial 
state along with a series of five observables in a three-dimensional Hilbert space. If so, 
then the quantum formalism can be used to model the effects of leading questions reported in the 
Ipsos data. 

On a closer inspection of the data, however, we find that the answer is negative. Note 
that asking a series of questions will result in a transformation of the state of mind represented 
by the density matrix. In particular, each time a question is asked, the state will transform 
in accordance with the von Neumann-L\"uders projection postulate \cite{Luders}, and there 
are consequences of this postulate. In particular, the resulting map acting on quantum state has to 
satisfy the complete positivity condition \cite{BH2015}. One implication here is that if $p=\max(p_i)$, 
where the set $\{p_i\}$ represents the probabilities of different outcomes in one question, 
and if $\{q_i\}$ are the probabilities of the various outcomes in the subsequent question, with 
$q=\max(q_i)$, then it is necessary that $q\leq p$. In other words, a quantum transformation 
can increase the degree of mixture, i.e. it can reduce the purity, but not otherwise. 

The simplest way of seeing this is to consider a pair of binary questions $F$ and $B$. We 
use the same notation as in the previous section, but $F$ and $B$ can be any pair of 
binary questions. Let the initial state be pure and is given by 
\[
|\psi\rangle = \sqrt{p}\, |F\rangle + \sqrt{1-p}\,|{\bar F}\rangle . 
\] 
The probability of detecting $F$ is therefore $p$, and that of detecting ${\bar F}=\neg F$ is 
$1-p$. The 
question is whether we can find an observable $B$ such that the subsequent measurement 
will yield an outcome with a probability that is larger than $\max(p,1-p)$. Now having 
performed the $F$ measurement, the initial state $|\psi\rangle$ transforms into 
\[ 
{\hat\rho}_F = p \, |F\rangle\langle F| + (1-p) \, |{\bar F}\rangle\langle{\bar F}| . 
\] 
For the $B$ measurement it suffices to consider a single projection operator $|B\rangle 
\langle B|$. Then the probability of obtaining a positive outcome in the $B$ measurement 
is 
\[ 
{\rm tr}({\hat\rho}_F|B\rangle \langle B|) = p\, |\langle B|F\rangle|^2 + (1-p) \, 
|\langle B|{\bar F}\rangle|^2 ,
\]
but writing $|\langle B|F\rangle|^2=r$ we have $|\langle B|{\bar F}\rangle|^2=1-r$, 
where $0\leq r\leq 1$. Hence we obtain 
\[ 
{\rm tr}({\hat\rho}_F|B\rangle \langle B|) = p\, r + (1-p) \, (1-r) , 
\]
which is a probabilistic average of $p$ and $1-p$, which lies in the interval between 
$p$ and $1-p$. It follows that there exists no observable $B$ such that the detection 
probability of either of its eigenvalues exceeding the probabilities of the outcomes of the 
previous $F$ measurement. 

This ``contraction'' property of quantum transformations persists in higher dimensions. For 
example, if $F$ is an observable representing a question with three answers $F_1$, $F_2$, 
and $F_3$, then writing the initial state in the form $|\psi\rangle=\sqrt{p_1}|F_1\rangle+
\sqrt{p_2}|F_2\rangle+\sqrt{p_3}|F_3\rangle$, the state after observing $F$ becomes 
\[ 
{\hat\rho}_F = p_1 \, |F_1\rangle\langle F_1| + p_2 \, |F_2\rangle\langle F_2| + 
p_3 \, |F_3\rangle\langle F_3| . 
\] 
The question thus is whether one can find a state $|B\rangle$ such that 
$\langle B|{\hat\rho}_F|B\rangle>\max(p_1,p_2,p_3)$. Because $|B\rangle$ is arbitrary, 
writing $|B\rangle=\sqrt{r_1}|F_1\rangle+\sqrt{r_2}|F_2\rangle+\sqrt{r_3}|F_3\rangle$, 
we see at once that 
\[
\langle B|{\hat\rho}_F|B\rangle = r_1 \, p_1 + r_2 \, p_2 + r_3 \, p_3 \, , 
\]
which again is a probabilistic average of $(p_1,p_2,p_3)$, and hence it cannot exceed 
$\max(p_1,p_2,p_3)$, or take a value below $\min(p_1,p_2,p_3)$ for that matter. 

One may relax the requirement of mutually orthogonal measurements 
and consider instead a positive operator-valued measure (POVM), which might in fact be 
more adequate to model general human behaviour because the repeatability of orthogonal 
measurements (that if a measurement is immediately followed by another measurement of 
the same observable then the outcome must be the same \cite{Luders}) 
need not hold in general: If a 
person is repeatedly asked the same question, sometimes the answer can flip. Indeed, 
Appleby's leading questions are all binary questions, and yet there are three answers to 
each. One can model this either by a three-dimensional orthogonal measurement, or by a 
three-component POVM in two dimensions. 
However, even if we allow for the use of an arbitrary POVM, the contraction property 
remains. Yet, looking at the data on Table \ref{Table:1} we observe, for instance, that 
going from the first question to the second question, the largest probability is increasing, 
and the smallest probability is decreasing; both violating the contraction property. It follows 
that the data shown in Tables \ref{Table:1} and \ref{Table:2} cannot be explained using 
quantum probability assignment rules, if the questions were to be modelled as observables 
acting on the same three-dimensional Hilbert space (or as POVMs on a two-dimensional 
Hilbert space).

\section*{Classical vs quantum probabilities} 

The foregoing formulation of using incompatible observables to represent Appleby's leading 
questions is not the only way to model them. In classical probability theory, if there are two 
questions, then they are modelled as two random variables that may be mutually dependent 
or not. In either case, there is no constraint resulting from the contraction property. For 
instance, if the two random variables are independent, then clearly the statistics for the 
first question will not have any implication on the second. The issue, however, is that the 
discrepancy of Question 5 in the two samples cannot be explained using classical 
probability theory. What I would like to address now therefore is whether a quantum 
analogue of a multi-variate probability distribution might be able to explain the data. 

To this end we remark that classical probability theory can be formulated on Hilbert space 
by means of the square-root embedding \cite{Rao,BH}. For instance, if $X$ is a binary 
random variable taking the values $(x_1,x_2)$ with probabilities $(p_1,p_2)$, then it can 
be modelled as a matrix with eigenvalues $(x_1,x_2)$ on Hilbert space, while the state is 
given by a vector with components $(\sqrt{p_1},\sqrt{p_2})$ in the basis of $X$. If there is 
a second binary random variable $Y$, then 
the state of the joint probability is represented as a normalized vector in the tensor 
product of a pair of two-dimensional Hilbert spaces. 
In particular, if the random variables are independent, then the state 
is a product state; whereas if they are dependent, then we have an entangled state. One 
might wonder if a state with a nontrivial entanglement property might explain 
the data shown in Tables \ref{Table:1} and \ref{Table:2}. 

To answer this question, it will be useful to clarify first the distinction between classical and 
quantum probabilities. To wit, there are two features that distinguish quantum probability 
from its classical counterpart: (a) observables (random variables) need not be compatible, 
and (b) probability amplitudes are complex numbers rather than real numbers. In the 
literature it is often stated, somewhat misleadingly, that entanglement is another feature 
that is unique to quantum theory, and this point of view is sometimes justified by the 
possible violation of the Bell inequality in quantum mechanics. It should be noted, however, 
that the various Bell inequalities are concerned with the statistics involving incompatible 
observables that cannot be treated using classical probability theory. The violation of the 
Bell inequality therefore is merely a statement of local realism \cite{Isham}, and not about 
a violation of classical probability rule because the latter theory makes no prediction about 
the relevant statistics. In other words, the bounds in the various Bell-type inequalities are 
obtained under the assumption of the existence of joint density functions for the variables 
involved \cite{Landau}, but such densities do not exist. 

Having noted this, one can then clarify the special status of entanglement, or more generally 
that of 
coherent superposition, in quantum theory. When classical probability theory is formulated 
on Hilbert space, a joint density of a pair of (or more) variables is mapped to an entangled 
vector. For a pair of binary variables, for instance, a perfectly anti-correlated classical state 
may take the form of a superposition $(|10\rangle+|01\rangle)/\sqrt{2}$, which happens to 
be a maximally entangled state (a measure of entanglement of a pure state is defined by the 
distance from the given state, on the state space, to the nearest disentangled state \cite{BH2}). 
However, if all observables are compatible, which is the 
case classically, then the statistics resulting from the use of entangled pure state $(|10\rangle+
|01\rangle)/\sqrt{2}$ are indistinguishable from those resulting from the corresponding mixed 
state $(|10\rangle\langle10|+|01\rangle\langle10|)/2$. Hence although the notion of a 
coherent superposition and, in particular, an entangled state, is in itself an artefact of the 
Hilbert space structure and statistical dependency in classical probability, its special status 
in quantum theory is a consequence of the fact that there are incompatible observables 
that enable the distinction between the statistics arising from the pure state $(|10\rangle+
|01\rangle)/\sqrt{2}$ and those arising from the mixed state $(|10\rangle\langle10|+|
01\rangle\langle10|)/2$. For the same token, entangled states may play a nontrivial role 
in modelling the cognitive state of mind, when considerations are made about incompatible 
propositions.

\section*{Can entanglement explain leading question effect?} 

With these considerations, let us represent the set of five questions in, say, Table 
\ref{Table:1}, as five local observables, each acting on its own Hilbert space, while the state 
of mind is modelled as an arbitrary (possibly entangled) state in the $3^5=243$-dimensional 
Hilbert space. Here, an observable that corresponds, say, to the second question $Q_2$ is 
said to be ``local'' if it takes the form $I_1\otimes Q_2\otimes I_3 \otimes I_4 \otimes I_5$, 
where $I_k$ denotes the identity matrix in the $k$th Hilbert space. 
In such a model, while the increase in the maximum probability (or decrease 
in the minimum probability) from one question to another can easily be explained. However, 
admissible quantum transformation rules satisfying the complete positivity condition impose 
a different constraint here to prevent the 
possibility of explaining the ``leading question'' effect. That is, based on the use of any set of 
local observables, the statistics of the outcome of the fifth question cannot be different 
between the two samples. 

In quantum theory this feature is sometimes referred to as the ``no-signalling'' property. To 
see this, consider a situation involving an entangled state of a physical quantum system, 
rather than a state of mind. Specifically, 
suppose that we were treating a system of five spin-1 particles prepared in an 
entangled state. We isolate one of them and separate it far away from the remaining four 
particles. Suppose further that many identical copies of such a quintic system are available. 
An observer X has the samples of the four particle systems, while observer Y has the 
samples of the remaining single-particle systems. At a pre-agreed time, observer X 
performs for each sample either Series A spin measurements in various directions, or 
Series B spin measurements in another set of directions, on the four particles. Observer Y 
knows the set of measurements in the two series, but does not know which set of 
measurements observer X has chosen. If, however, by analyzing the statistics of a 
measurement performed on the fifth particle, observer Y can determine which one of the 
two series of experiments observer X has performed, then one bit of information has 
been transmitted. Because the physical distance between X and Y can be arbitrarily large, 
it follows that this setup would permit an arbitrarily fast information transmission. 
Transformations of quantum states that satisfy the complete positivity conditions, such as 
a series of measurements, however, excludes such a possibility.  

When one models the cognitive state of mind, the notion of nonlocality plays essentially no role, 
since we do not make decisions while part of our brain is being transported to the moon. 
Nevertheless, quantum cognition relies on the same mathematical structures as quantum 
mechanics for probability assignments, and hence the no-signalling rule is applicable 
to show that leading-question effect cannot be explained by using a set of local observables 
to model the questions, irrespective of what kind of entangled state one might have.

\section*{Ipsos data on Appleby's questions can be modelled by isolating the first question} 

Another approach that in principle can work to model the Ipsos data is to regard, for 
each sample, the 
first question as an independent isolated question, modelled on its own 3-dimensional 
Hilbert space. The remaining four questions are then modelled on yet another 3-dimensional 
Hilbert space as four incompatible observables. In other words, it is analogous to having 
two particles in a disentangled state for which on the first particle one measurement is 
performed, while on the second particle a series of four incompatible measurements is 
performed. Because the state is disentangled, the outcome of the measurement on the 
first particle will impose no constraint on the measurements on the second particle.  
(In Sample B, the map 
going from the second to the third question also violates the complete positivity, but the 
violation here is relatively small, so one could reasonably argue that it falls within the error 
margin.) That is, we express the initial state as a product state 
\[ 
|\psi\rangle = \left( \begin{array}{c} p_1 \\ p_2 \\ p_3 \end{array} \right) \otimes 
\left( \begin{array}{c} q_1 \\ q_2 \\ q_3 \end{array} \right) 
\]
in a nine-dimensional Hilbert space, where the probabilities $\{p_i\}$ refer to those of the 
first question and $\{q_i\}$ to the second equation, when the state is expressed in the 
basis of these two questions. 

To see how the Ipsos data can be modelled in this scheme, it suffices to show how it is 
possible to go from one question to the next in the remaining four questions. Once the 
second question is asked, this will fix the density matrix: 
\[ 
{\hat\rho}_2 = q_1 \, |Q_2^1\rangle\langle Q_2^1| + q_2 \, |Q_2^2\rangle\langle Q_2^2| + 
q_3 \, |Q_2^3\rangle\langle Q_2^3| ,
\] 
when expressed in the basis of the second-question observable. Note that here I am 
focusing on the three-dimensional Hilbert subspace associated to the latter four questions. 

Assuming that the third question does not violate the contraction constraint, what we 
need is to find a triplet of orthogonal directions on the three-dimensional Hilbert space 
such that the expectations of the density matrix in the three directions give the probability 
for the next question. That is, writing $(r_1,r_2,r_3)$ for the probabilities of the third 
question, we need to find a triplet of orthogonal states 
$(|Q_3^1\rangle,|Q_3^2\rangle,|Q_3^3\rangle)$ such that 
\[ 
\langle Q_3^1|{\hat\rho}_2|Q_3^1\rangle = r_1, \quad 
\langle Q_3^2|{\hat\rho}_2|Q_3^2\rangle = r_2, \quad 
\langle Q_3^3|{\hat\rho}_2|Q_3^3\rangle = r_3.
\]
On account of the normalization, however, once the first two conditions are met, the 
third one is automatically satisfied. Hence these conditions give just two constraints. 

Now there are six real parameters -- three angles and three phases -- to specify a triplet 
of orthogonal directions $(|Q_3^1\rangle,|Q_3^2\rangle,|Q_3^3\rangle)$ on a 
three-dimensional Hilbert space (see \cite{BEH} on how orthogonal frames in Hilbert 
space can be parametrized). However, two of the phase variables will drop out of the 
equations. Hence we are left with four free parameters to fit two constraints, which in 
general can be achieved. Once the third question is asked, the state now transforms to 
\[ 
{\hat\rho}_3 = r_1 \, |Q_3^1\rangle\langle Q_3^1| + r_2 \, |Q_3^2\rangle\langle Q_3^2| + 
r_3 \, |Q_3^3\rangle\langle Q_3^3| ,
\] 
and we just have to find a new triplet $(|Q_4^1\rangle,|Q_4^2\rangle,|Q_4^3\rangle)$ 
following the same procedure, and so on for the subsequent question. 

The scheme proposed here thus can in principle explain the data. However, the validity 
of isolating the first question can only be justified by a new experiment. Namely, if the 
same set of questions, but without the first one, are asked and if the percentages of 
answers are consistent with those of the remaining four questions provided in Tables 
\ref{Table:1} and \ref{Table:2}, then it is legitimate to regard the first question as being 
independent. On the other hand, if the result substantially changes the data, then this 
scheme cannot be justified.

\section*{Discussion} 

We have shown that in general the leading-question effect, while cannot be explained using 
classical 
probability theory, can nonetheless be explained using quantum probability theory. However, 
there are constraints within the allowable quantum transformations that may prevent the 
modelling of certain data. In particular, when it comes to the Ipsos data presented in Tables 
\ref{Table:1} and \ref{Table:2}, they cannot be modelled using the most straightforward 
approachs in terms of a set of incompatible ``global'' observables, or in terms of a set of 
local observables in an entangled state. A workable scheme has been provided, but the 
approach requires a justification in the form of further experimental data. 

Should this scheme fail the experimental test, there still remain other possibilities. For 
example, consider a 
question having three or more answers, modelled on a Hilbert space of dimension three 
or larger. Having settled this question, if one were to ask a ``degenerate'' binary question, 
then in this case the probability of an affirmative answer can exceed any of the probabilities 
of the first question, without violating the contraction constraint. Such a model, however, 
does not lends itself with the context of Appleby's questions considered here. Nevertheless, 
it may be possible that by a clever construction involving a large-dimensional Hilbert space 
along with partially nonlocal degenerate observables and an entangled initial state, one 
can model the Ipsos data using quantum probability rules. Whether such a construction 
exists to fit the data of Tables \ref{Table:1} and \ref{Table:2} is an open question. 

In traditional cognitive psychology, when the behaviour of people in their reasoning and decision making 
does not follow the rules of classical 
probability theory, then they are classified as being ``irrational" \cite{Evans}. 
Quantum cognition expands 
the scope of mathematics underpinning outcome of chance used in classical probability 
theory to re-classify many examples of such ``irrational'' behaviour as being perfectly rational, because 
they can be modelled consistently using quantum probability theory. An important 
ingredient in the use of quantum probability in modelling cognitive science, 
apart from the existence of incompatible propositions, is how the state of 
mind transforms in reasoning and decision making. Typically, this is governed 
by the von Neumann-L\"uders rule \cite{Luders}. Now this rule has several advantages. To 
begin, if all proposition or information being processed are compatible, then the rule 
reproduces the Bayesian 
logic \cite{Ozawa,Brody2023}. This follows from the fact that conditioning in classical 
probability merely implies projection on Hilbert space. It provides a way of ``separating 
what we know from what we want to know'' \cite{Reid} in the most transparent manner. 
Hence the von Neumann-L\"uders rule encompasses the classical Bayesian thinking, but it 
can also explain wider phenomena. More 
importantly, however, in processing information about the external world, this rule provides 
an assessment of the uncertain outside world in a given circumstance that on average 
minimizes uncertainties, and that on average leads to minimum energy consumption in 
response to outside information \cite{Brody1}. Given these properties, it seems reasonable to 
postulate that cognitive behaviours that are consistent with the von Neumann-L\"uders rule 
should be biologically preferred, and hence be regarded as rational. Whether the Ipsos 
data on Appleby's questions represent a rational logic, however, remains an open question.

\vspace{0.3cm}
\begin{footnotesize}
\noindent {\bf Acknowledgements}. The author thanks Michael Kearney for drawing his 
attention to the episode of \textit{Yes, Prime Minister}, and Eva-Maria Graefe and Lane 
P. Hughston for stimulating discussion and comments. 
\end{footnotesize}

\end{document}